\pdfoutput=1

\documentclass[11pt]{article}

\usepackage{booktabs}
\usepackage{caption}
\usepackage{graphicx}
\usepackage{multirow}
\usepackage{tabularx}
\usepackage{adjustbox}
\usepackage{tablefootnote}
\usepackage{footnote}
\usepackage{threeparttable}
\usepackage{hyperref}

\usepackage[most]{tcolorbox} 
\usepackage{fontawesome5} 
\usepackage{enumitem} 

\usepackage[final]{coling}

\usepackage{times}
\usepackage{latexsym}

\usepackage[T1]{fontenc}

\usepackage[utf8]{inputenc}

\usepackage{microtype}

\usepackage{inconsolata}

\usepackage{graphicx}

\usepackage{xcolor}
\definecolor{WarningBrown}{rgb}{0.55, 0.27, 0.08} 
\definecolor{WarningReddishBrown}{rgb}{0.65, 0.25, 0.12}
\definecolor{WarningSienna}{rgb}{0.7, 0.2, 0.1}

\usepackage{colortbl}       
\definecolor{highlightgreen}{rgb}{0.9, 1.0, 0.9}

\title{Benchmarking Hindi LLMs: A New Suite of Datasets and a Comparative Analysis}

\author{Anusha Kamath, Kanishk Singla, Rakesh Paul, Raviraj Joshi,\\ \textbf{ Utkarsh Vaidya, Sanjay Singh Chauhan, Niranjan Wartikar} \\
        NVIDIA \\
        \texttt{\{anushak, kanishks, rapaul, ravirajj, uvaidya,} \\ \texttt{schauhan, nwartikar\}@nvidia.com}}

\begin{document}
\maketitle
\begin{abstract}

Evaluating instruction-tuned Large Language Models (LLMs) in Hindi is challenging due to a lack of high-quality benchmarks, as direct translation of English datasets fails to capture crucial linguistic and cultural nuances. To address this, we introduce a suite of five Hindi LLM evaluation datasets: IFEval-Hi, MT-Bench-Hi, GSM8K-Hi, ChatRAG-Hi, and BFCL-Hi. These were created using a methodology that combines from-scratch human annotation with a translate-and-verify process. We leverage this suite to conduct an extensive benchmarking of open-source LLMs supporting Hindi, providing a detailed comparative analysis of their current capabilities. Our curation process also serves as a replicable methodology for developing benchmarks in other low-resource languages.

\end{abstract}

\section{Introduction}
The rapid expansion of Large Language Models (LLMs) necessitates the development of robust and reliable evaluation methodologies \cite{liang2022holistic,srivastava2023beyond}. As these models are integrated into a wide range of applications, a rigorous assessment of their capabilities, limitations, and safety is paramount \cite{achiam2023gpt,wang2023decodingtrust}. Although the initial focus of evaluation has been predominantly on English, a model's global utility is contingent upon its performance across diverse linguistic and cultural contexts \cite{singh2024aya}. The evaluation of non-English LLMs is therefore essential, not only for ensuring equitable technological access but also for understanding the extent to which these models capture the distinct complexities inherent in different languages, an undertaking that goes beyond mere translation \cite{bender2021dangers}.

The evaluation landscape for English LLMs is well-established, featuring a comprehensive suite of benchmarks targeting a spectrum of model capabilities. For foundational "base" models, benchmarks assess commonsense reasoning, such as HellaSwag \cite{zellers2019hellaswag} and Winogrande \cite{sakaguchi2021winogrande}, factual accuracy with TruthfulQA \cite{lin2022truthfulqa}, and broad multi-task knowledge with MMLU \cite{hendrycks2020measuring,wang2024mmlu,singh2024global}. Specialized datasets evaluate capabilities like mathematical reasoning on GSM8K\cite{cobbe2021training} and code generation with HumanEval \cite{chen2021evaluating} and MBPP \cite{austin2021program}. Furthermore, the advent of interactive, instruction-following models has spurred the creation of benchmarks to assess conversational quality on MT-Bench \cite{zheng2023judging}, fidelity to complex instructions with IFEval \cite{zhou2023instruction}, and the ability to execute tool or function calls correctly on BFCL \cite{patilberkeley}. These datasets have collectively become the standard for evaluating the performance of state-of-the-art models in English.

\begin{table*}[t]
\centering
\begin{adjustbox}{width=\textwidth}
\begin{tabular}{lcl}
\toprule
\textbf{Dataset Name} & \textbf{Count} & \textbf{Method} \\
\midrule
IFEval-Hi & 848 & In-house \\
MT-Bench-Hi & 200 & Translated and human evaluated (4 categories); In-house (4 categories) \\
GSM8K-Hi & 1319 & Translated and human evaluated (100\%) \\
ChatRAG-Hi & 5948 & 
Translated, filtered, and human-evaluated (5\%). Includes: INSCIT (450), Doc2Dial (498), QuAC, QReCC, \\
& & TopiocQA, CoQA, HybriDial, SQA, DoQA (Cooking, Travel, Movies), ConvFinQA (500 each). \\
& & Context: GCP translated, no filtering. \\
& & Answers and conversation turns: \\
& & - Used GCP translated data when the back-translated version matched the original (CHRF++ $\geq$ 90). \\
& & - Else, used LLM translated data with heuristic filtering to remove poor translations. \\
BFCL-Hi & 2251 & Translated (not human evaluated) \\
\bottomrule
\end{tabular}
\end{adjustbox}
\caption{Overview of the Hindi evaluation datasets. The test suite consists of Hindi versions of IFEval, MT-Bench, GSM8K, ChatRAG, and BFCL.}
\label{tab:evaluation_dataset_details}
\end{table*}

In recent years, significant progress has been made in developing evaluation resources for Indic languages, typified by benchmarks such as IndicGLUE \cite{kakwani2020indicnlpsuite}, MILU \cite{verma2025milu}, IndicMMLU-Pro \cite{sankalp2025indicmmlu}, and IndicGenBench \cite{singh2024indicgenbench}. These resources have been instrumental in assessing the core capabilities of pre-trained base models across numerous languages of the Indian subcontinent \cite{joshi2024adapting}. Despite this progress, the existing benchmarks primarily target pre-trained base models, leaving a noticeable gap in resources for assessing the capabilities of instruction-tuned models. Consequently, benchmarks for critical skills like instruction following, conversational ability, and function calling, such as Hindi versions of IFEval, MT-Bench, and BFCL, are largely unavailable publicly.

A common methodology to address this gap involves the direct translation of existing English benchmarks. This approach, however, presents considerable challenges, as automated translation frequently fails to preserve the linguistic subtleties and cultural context integral to the target language. This process can yield datasets that are linguistically incongruous or culturally irrelevant, thereby diminishing the validity and reliability of the evaluation. Such benchmarks often test a model's ability to comprehend translated English rather than its native fluency and instruction fidelity.

To address these deficiencies, this paper introduces Hindi versions of five widely-used and comprehensive benchmarks: IFEval-Hi, MT-Bench-Hi, GSM8K-Hi, ChatRAG-Hi, and BFCL-Hi. We developed these datasets using a process that combined direct human creation with a translate-and-verify workflow, ensuring high linguistic and cultural relevance. A summary of the final dataset sizes and curation methods is presented in Table \ref{tab:evaluation_dataset_details}. Furthermore, we utilize this new suite to conduct a comprehensive benchmarking of several prominent, publicly available LLMs based on foundational models, including Llama, Gemma, and Nemotron. This work contributes a valuable, high-quality evaluation suite for Hindi to the research community and presents a comparative analysis that offers critical insights into the current capabilities of Hindi language models.

The main contributions of our work are as follows:
\begin{itemize}
    \item We introduce a suite of five new, high-quality benchmarks (IFEval-Hi\footnote{\href{https://huggingface.co/datasets/nvidia/IFEval-Hi}{https://huggingface.co/datasets/nvidia/IFEval-Hi}}, MT-Bench-Hi\footnote{\href{https://huggingface.co/datasets/nvidia/MT-Bench-Hi}{https://huggingface.co/datasets/nvidia/MT-Bench-Hi}}, GSM8K-Hi\footnote{\href{https://huggingface.co/datasets/nvidia/GSM8K-Hi}{https://huggingface.co/datasets/nvidia/GSM8K-Hi}}, ChatRAG-Hi\footnote{\href{https://huggingface.co/datasets/nvidia/ChatRAG-Hi}{https://huggingface.co/datasets/nvidia/ChatRAG-Hi}}, and BFCL-Hi\footnote{\href{https://huggingface.co/datasets/nvidia/BFCL-Hi}{https://huggingface.co/datasets/nvidia/BFCL-Hi}}) for evaluating instruction-tuned LLMs in Hindi and detail the curation process developed for their creation.
    \item We present a comprehensive benchmark of prominent, publicly available LLMs on this new suite, providing the first robust comparative analysis of their capabilities in Hindi. Our findings show that while specialized models exhibit strength in specific tasks, Gemma-2-9b-it in the SLM class and GPT-OSS-120B in the LLM class emerge as the most capable general-purpose models.
\end{itemize}

\section{Related Work}
Recent years have witnessed notable progress in the evaluation of multilingual and low-resource language models, with a particular focus on Indic languages. Foundational efforts, such as IndicGLUE \cite{kakwani2020indicnlpsuite} and IndicXTREME \cite{doddapaneni2023towards}, established the initial groundwork by adapting the GLUE paradigm for major Indic languages. These benchmarks provided a broad suite of Natural Language Understanding (NLU) tasks, including classification, entailment, and named entity recognition, which proved instrumental in assessing the foundational capabilities of models across multiple Indic languages, including Hindi.

Building upon these foundations, subsequent benchmarks like MILU \cite{verma2025milu} introduced more challenging and culturally grounded tasks. MILU, is a large-scale benchmark comprising approximately 80,000 multiple-choice questions derived from Indian competitive examinations. By emphasizing India-specific domains such as local governance, arts, and history, MILU underscores the importance of cultural context in evaluation, an element often diluted in directly translated datasets. Specialized datasets like IndicQuest \cite{rohera2024l3cube} have been developed to evaluate the factual knowledge of Indic LLMs. In parallel, benchmarks such as IndicSQuAD \cite{endait2025indicsquad} and IndicQA \cite{singh2025indic} have addressed extractive and abstractive question answering.

More recently, the field has shifted toward multi-task and generative evaluation. Benchmarks like the IndicGenBench suite \cite{singh2024indicgenbench} and IndicMMLU-Pro \cite{sankalp2025indicmmlu} now assess complex reasoning, creative understanding, and instruction-following, demonstrating a move beyond traditional NLU paradigms. This trend is further reflected in the Okapi \cite{lai2023okapi}, which translated key English benchmarks into numerous languages, and the development of Global MMLU \cite{singh2024global}, which extends evaluation to more diverse cultural contexts.

\begin{figure}[t]  
    \centering
    \includegraphics[width=\columnwidth]{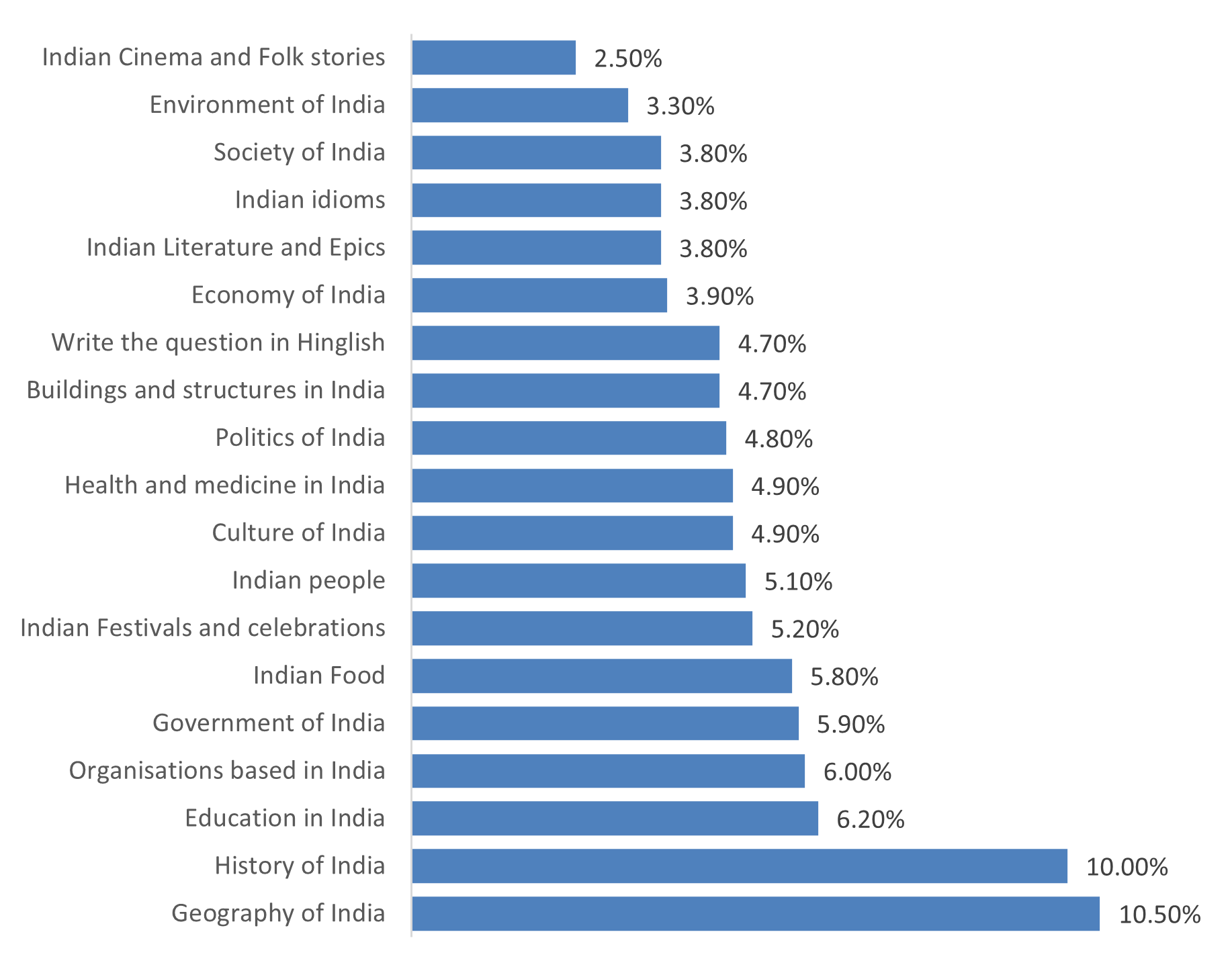} 
    \caption{Distribution of samples by Indian cultural themes in the IFEval-Hi dataset.}
    \label{fig:ifeval_domain}
\end{figure}

\begin{figure}[t]  
    \centering
    \includegraphics[width=\columnwidth]{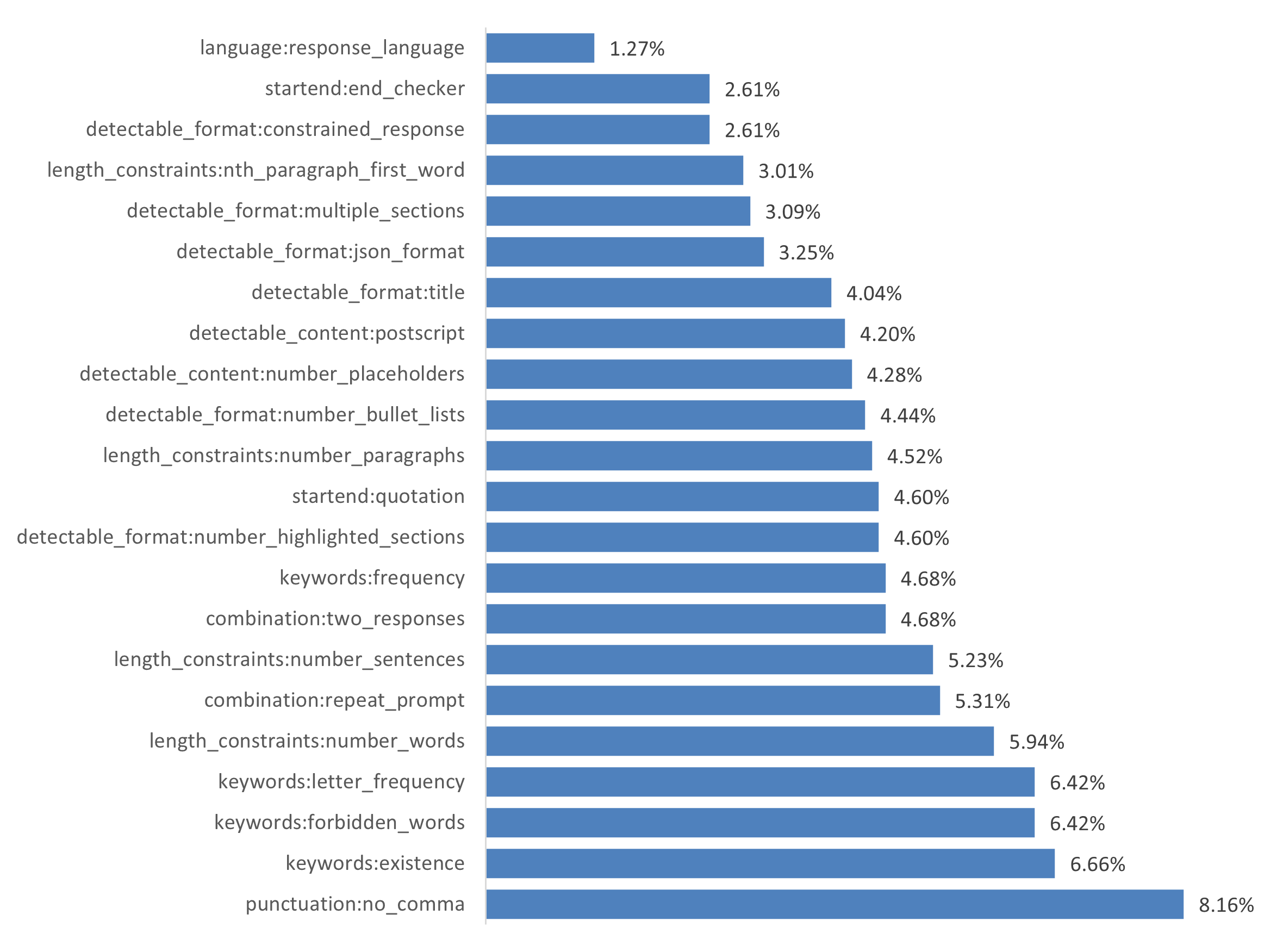}
    \caption{Distribution of verifiable instruction types within the IFEval-Hi dataset.}
    \label{fig:ifeval_instructions}
\end{figure}

\begin{figure}[t]  
    \centering
    \includegraphics[width=\columnwidth]{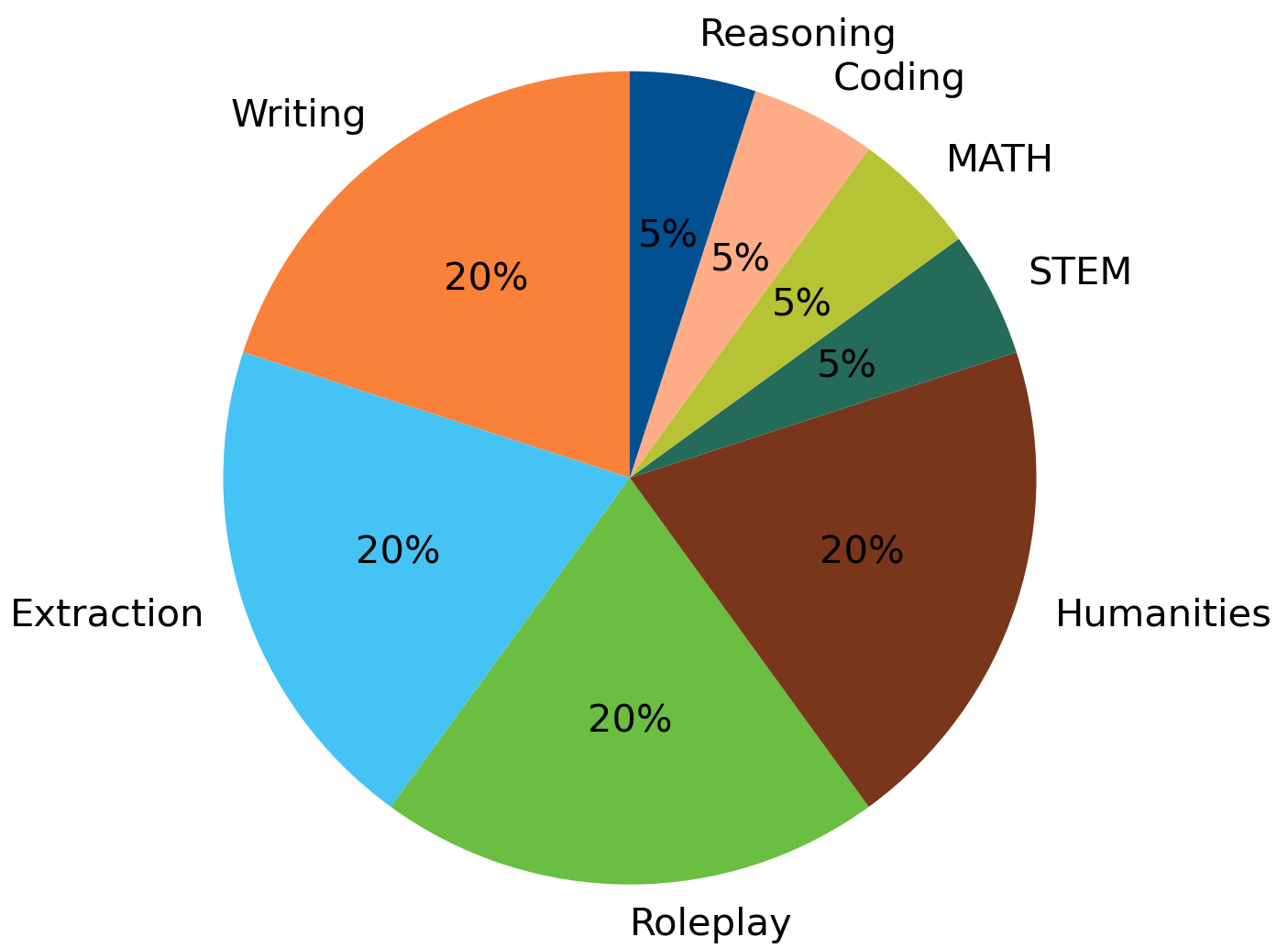}  
    \caption{Category distribution in MT-Bench-Hi, adapted with Indian cultural themes to increase focus on culturally relevant instructions.}
    \label{fig:mtbench_categories}
\end{figure}

\section{Dataset Curation}

\begin{figure*}[h]  
    \centering
    \includegraphics[scale=0.7]{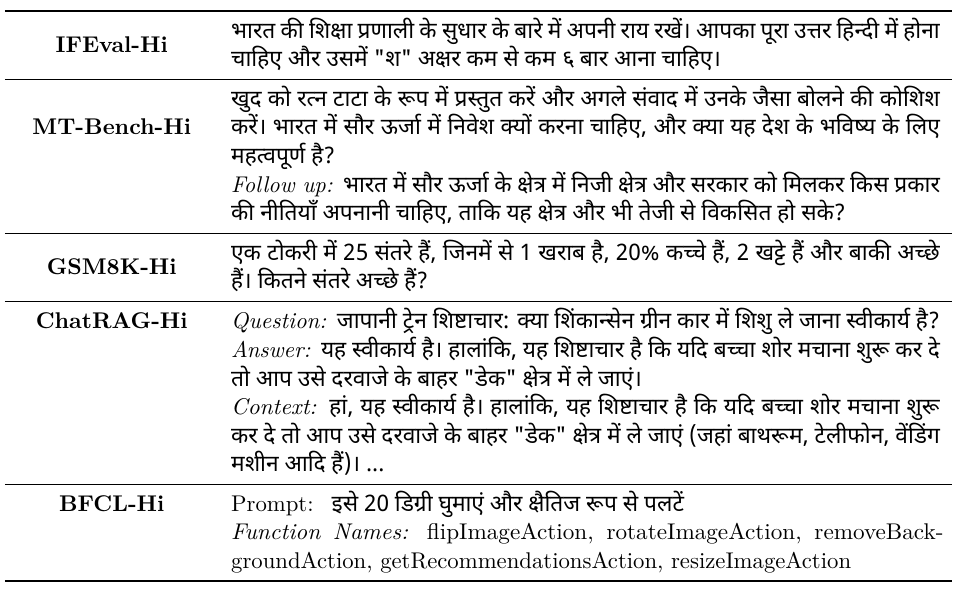}
    \caption{Representative examples from five Hindi evaluation datasets curated in this study.}
    \label{fig:all_hi_samples}
\end{figure*}

This section describes the process used to curate the Hindi versions of popular English benchmark datasets. Sample examples from each dataset are shown in Figure \ref{fig:all_hi_samples}.
\subsection{IFEval-Hi}
The creation of IFEval-Hi is based on the English Instruction Following Evaluation (IFEval) benchmark, which is designed to rigorously assess an LLM's ability to adhere to precise instructions. The original English IFEval is structured around 25 distinct and objectively verifiable instruction types, such as "insert a word at a specific position" or "reverse the first paragraph". This structure ensures a reliable and scalable evaluation process. To test models with increasing difficulty, the prompts are organized into three complexity levels: Single, Double, and Triple Instructions, requiring the model to execute one, two, or three distinct commands within the same prompt, respectively.

Our curation process for IFEval-Hi was a systematic adaptation of this framework to an Indian cultural and linguistic context. The core of this process involved retaining the 22 verifiable instruction types as a structural framework while replacing the generic content of the English prompts with themes relevant to India. Some categories that are not relevant to Hindi, such as "Change Cases," were dropped. The thematic content was sourced from comprehensive categories on Wikipedia related to India, covering a wide range of topics including Indian history, philosophy, festivals, art forms, and social norms. The distribution of cultural themes is detailed in Figure \ref{fig:ifeval_domain}, while the breakdown across verifiable instruction categories is presented in Figure \ref{fig:ifeval_instructions}.

New prompts were carefully created by a team of five annotators over a ten-week period. To ensure that the newly created Indian-themed prompts were both culturally relevant and objectively verifiable, annotators were provided with examples for each instruction type. For instance, when the instruction theme is "Geography of India" and the instruction category is a letter frequency constraint, such as requiring a certain Hindi letter to appear at least three times, the annotator crafts an instruction that incorporates both the theme and the explicit constraint. Specific sample is shown in Figure \ref{fig:all_hi_samples}. To ensure that IFEval-Hi could be used as a direct benchmark against its English counterpart, the evaluation metrics and constraints for each of the 22 instruction categories were directly mirrored, along with the three levels of complexity. This significant human-in-the-loop effort resulted in a final dataset comprising 848 high-quality, culturally resonant samples. The annotation process is described in further detail in Appendix \ref{subsec:appendix_ifeval}.

\begin{figure*}[h]  
    \centering
    \includegraphics[scale=0.45]{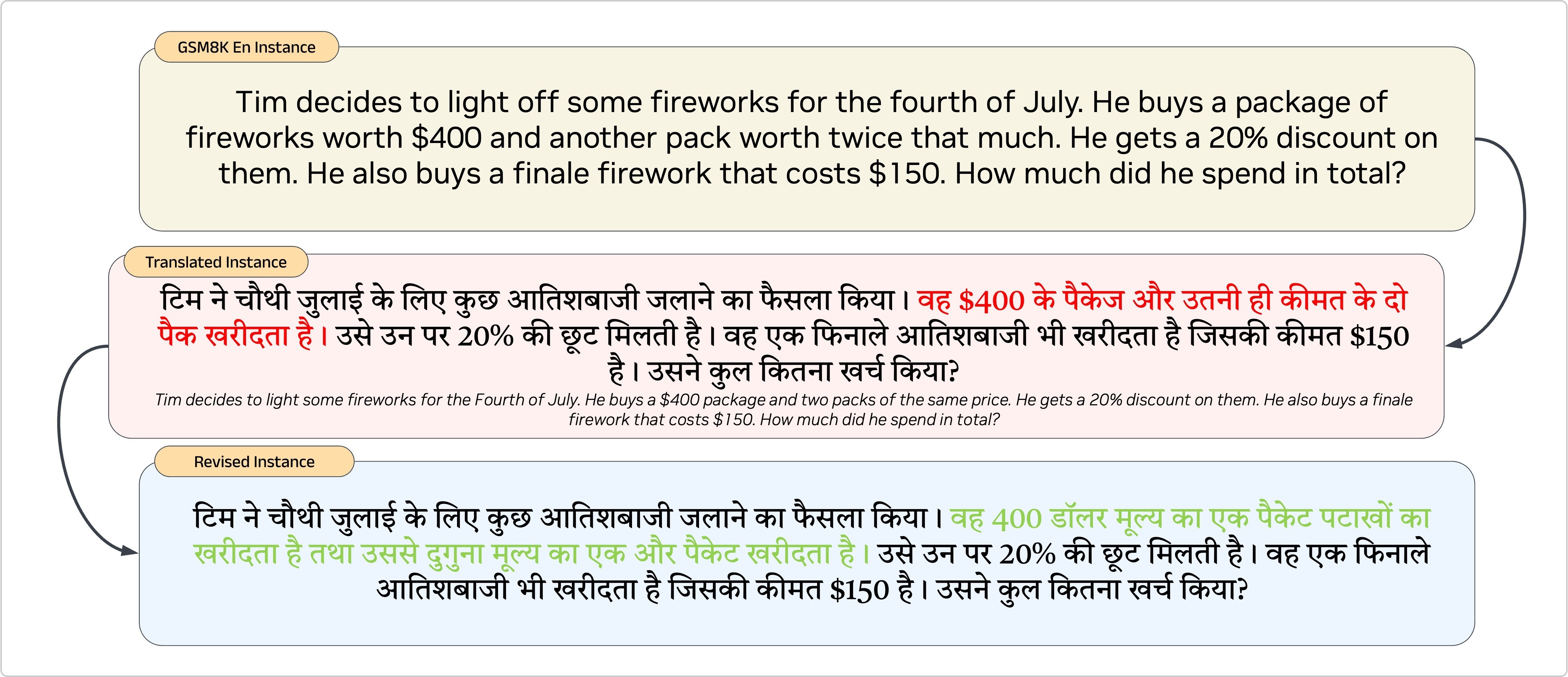}
    \caption{A sample GSM8K question highlighting a translation mistake in Hindi (red), the corrected version (green), and the corresponding English line (yellow), showcasing the process of identifying and fixing language conversion errors manually.}
    \label{fig:gsm8k-sample}
\end{figure*}

\subsection{MT-Bench-Hi}
MT-Bench-Hi is the Hindi adaptation of the English Multi-Turn Benchmark (MT-Bench), a standard for evaluating the conversational and reasoning abilities of LLMs in extended dialogues. The original benchmark consists of 80 high-quality, multi-turn questions designed to test key capabilities such as maintaining context, response accuracy, and instruction following. It employs an "LLM-as-a-Judge" approach, where a powerful model like GPT-4o scores all responses on a 1-10 scale using two distinct methods: for reference-free categories (STEM, Writing, Roleplay, Humanities, Extraction), responses are scored directly, while for categories with reference answers (Reasoning, Math, Coding), they are evaluated via pairwise comparison against the reference answer.

The curation of MT-Bench-Hi was a detailed adaptation process designed to make the benchmark culturally and contextually relevant for India. We adopted a hybrid approach to content creation. For universal technical categories (STEM, math, reasoning, coding), questions were translated from English to Hindi using GCP and subsequently underwent thorough human evaluation to verify accuracy and intent. For categories requiring deep cultural contextualization (Writing, Roleplay, Humanities, Extraction), questions were created from scratch by human specialists to ensure the prompts were authentically Indian. Figure \ref{fig:mtbench_categories} illustrates the final distribution of samples across these categories.

To maintain high standards, annotators were provided with reference examples from the English MT-Bench and guided through a specialized interface. A key quality assurance step involved showing annotators sample responses from a high-performing model (e.g., GPT-4o) to help them craft prompts that could effectively test advanced capabilities in an Indian context. The evaluation framework was aligned with the original's "LLM-as-a-Judge" methodology. To ensure consistency, we maintain the same format as the original dataset; for categories that include a reference answer, we retain the original English reference answer during evaluation. Aligning with the original MT-Bench setup, we employ direct, single-answer evaluation for reference-free, subjective categories (e.g., Writing, Roleplay) and pairwise comparison against a reference answer for categories with objective solutions (e.g., Math, Coding, Reasoning).
The annotation process is described in further detail in Appendix \ref{subsec:appendix_mtbench}.

\subsection{GSM8K-Hi}
The foundation for this dataset is the English GSM8K (Grade School Math 8K), a prominent benchmark for assessing the mathematical reasoning of LLMs. Directly translating the dataset into Hindi risks altering the underlying mathematical logic, particularly in problems with comparative constructs such as ‘twice that amount’ or ‘10 less than half the age of’. However, crafting linguistically diverse math problems that require multi-step solutions demands significant expertise in both mathematics and language structure. This process involves considerable time and effort from human annotators with domain expertise, making it a resource-intensive endeavor. Therefore, to balance quality with feasibility, we opted for a two-step "translate-then-verify" methodology.

The process began with machine translation of the original English problems (including GSM8K system prompt) into Hindi using GCP. Human annotators were then provided with both the Hindi translation and the original English text for reference. Their primary task was to evaluate the Hindi translation for correctness and suggest modifications to ensure linguistic accuracy and contextual appropriateness. This verification stage proved to be essential, as annotators flagged approximately 10\% of the machine-translated data for inaccuracies. These instances were subsequently reviewed and corrected through close collaboration between our development team and the annotators, ensuring the final dataset maintained high quality. To illustrate this process, Figure \ref{fig:gsm8k-sample} shows a sample error alongside its correction. To ensure consistent benchmarking, GSM8K-Hi is evaluated using the LM-Eval-Harness, the same framework employed for the original English dataset. The annotation process is described in further detail in Appendix \ref{subsec:appendix_gsm8k}.


\begin{table*}[ht]
\centering
\begin{adjustbox}{scale=0.7}
\begin{tabular}{lcccccc}
\toprule
\textbf{Model} & \textbf{Size} & \textbf{MT-Bench-Hi} & \textbf{BFCL-Hi} & \textbf{GSM8K-Hi} & \textbf{IFEval-Hi} & \textbf{ChatRAG-Hi} \\
\midrule
\multicolumn{7}{c}{\textbf{SLMs}} \\
\midrule
Gemma-2-2b-it & 2B & 4.37 & 32.96 & 26.99 & 38.92 & 29.89 \\
Llama-3.2-3B-Instruct & 3B & 5.14 & 33.81 & 40.11 & 40.80 & 32.60 \\
Nemotron-Mini-4B-Instruct & 4B & 3.44 & - & 32.22 & 36.08 & 27.32 \\
Nemotron-4-Mini-Hindi-4B-Instruct & 4B & 6.01 & \textbf{52.82} & 47.31 & 51.65 & 36.07 \\
Llama-3.1-8B-Instruct & 8B & 6.44 & 31.23 & 61.33 & 48.82 & 38.03 \\
Aya-expanse-8b & 8B & 6.58 & 36.56 & \textbf{64.52} & 42.92 & 30.15 \\
\rowcolor{highlightgreen} 
Gemma-2-9b-it & 9B & \textbf{7.37} & 50.51 & 64.44 & \textbf{61.79} & \textbf{40.97} \\
Krutrim-2-instruct & 12B & 6.31 & 26.88 & 56.56 & 59.32 & 37.48 \\
\midrule
\multicolumn{7}{c}{\textbf{LLMs}} \\
\midrule
GPT-OSS-20B (reasoning low) & 21B & 8.51 & 54.60 & 80.64 & 69.04 & 26.16 \\
Mistral-Small-3.2-24B-Instruct-2506 & 24B & 7.83 & 41.45 & 77.55 & 66.89 & 37.92 \\
Sarvam-M (reasoning off) & 24B & 8.25 & 48.60 & 82.30 & 71.64 & 40.14 \\

Gemma-3-27b-it & 27B & 8.31 & \textbf{62.42} & 78.12 & 67.72 & 45.23 \\
\rowcolor{highlightgreen} 
GPT-OSS-120B (reasoning low) & 117B & \textbf{8.70} & 61.26 & \textbf{93.41} & \textbf{73.86} & 29.85 \\
Qwen3-235B-A22B-FP8 (reasoning off) & 235B & 8.10 & 59.88 & 89.69 & 68.11 & 32.47 \\
Llama-3.1-405B & 405B & 7.17 & 49.53 & 86.27 & 68.66 & \textbf{47.46} \\
\midrule
\multicolumn{7}{c}{\textbf{LLMs (Reasoning)}} \\
\midrule
GPT-OSS-20B (reasoning medium) & 21B & 8.43 & 63.26 & 83.41 & 72.01 & 29.16 \\
GPT-OSS-20B (reasoning high) & 21B & 8.23 & 64.77 & 83.44 & 72.11 & 32.39 \\
Sarvam-M (reasoning on) & 24B & 8.60 & 59.53 & 84.40 & 74.06 & \textbf{37.13} \\
\rowcolor{highlightgreen} 
GPT-OSS-120B (reasoning medium) & 117B & \textbf{8.79} & \textbf{66.19} & 95.93 & 76.69 & 30.80 \\
GPT-OSS-120B (reasoning high) & 117B & 8.70 & 64.90 & \textbf{96.27} & \textbf{76.80} & 31.82 \\
\bottomrule
\end{tabular}
\end{adjustbox}
\caption{Performance of various LLMs on Hindi benchmarks. MT-Bench-Hi is scored on a scale of 1-10 using an LLM-as-a-judge approach. BFCL-Hi, GSM8K-Hi, and IFEval-Hi report accuracy on a 1-100 scale, while ChatRAG-Hi reports the F1-Score. The highest score in each column is highlighted in bold.}
\label{tab:hindi_model_benchmarks}
\end{table*}

\subsection{ChatRAG-Hi}
ChatRAG-Hi is the Hindi version of ChatRAG Bench, a benchmark for evaluating conversational question-answering using documents and retrieved context. The original incorporates ten diverse datasets, including Doc2Dial, QuAC, and ConvFinQA. Adapting this composite dataset posed challenges due to the varied structures of its subsets, which range from extensive contexts to single-word answers.

Our curation process involved a differential translation strategy. The extensive context passages were translated using GCP without subsequent filtering. For the more sensitive answers and conversation turns, we adopted a two-tiered approach. We first used GCP and validated the output by back-translating it to English. If the back-translated text matched the original with a high degree of fidelity (CHRF++ score >= 90), the GCP translation was retained. In cases where the CHRF++ score was low, which often occurred with very short text segments (1–3 words) where GCP lacks sufficient contextual cues, the GCP translation was discarded. To overcome this, we employed an LLM for these segments, providing it with the broader GCP-translated Hindi context alongside the original short English answer to generate a more accurate and contextually appropriate Hindi equivalent. This LLM-generated (Llama-3.1-405B) data was then subjected to heuristic filtering to remove poor-quality outputs. This hybrid methodology was designed to maximize accuracy across different text types. To ensure overall quality, approximately 10\% of the final Hindi data underwent human verification, which confirmed the high fidelity of the translations, with the error rate across subsets remaining within 1-5\%.

\subsection{BFCL-Hi}
BFCL-Hi is the Hindi adaptation of the Berkeley Function-Calling Leaderboard (BFCL V2), a benchmark designed to evaluate the ability of LLMs to call functions or tools. The original dataset comprises diverse function-calling scenarios, including simple, multiple, and parallel calls. It also includes relevance and irrelevance categories to assess a model's ability to determine if the provided tools are appropriate for a given query.

The dataset is structured in a JSON format where each entry contains a conversation history and an array of available functions, defined with names, descriptions, and parameter schemas. To create BFCL-Hi, we translated the conversational history into Hindi using the GCP translation service. Crucially, the function calls themselves, including their names, descriptions, and parameter details, were retained in their original English format. This hybrid approach tests the model's ability to understand a Hindi query and map it to a predefined English-language tool. However, to make the dataset more relevant for fully localized use cases, the function parameters should also be translated into Hindi, which we leave as a task for future work. The ground truth for simple, multiple, and parallel categories remained unchanged from the English version. The relevance and irrelevance categories do not include ground truth, as they are designed to verify whether the model correctly attempts a function call. Evaluation is performed using the BFCL Abstract Syntax Tree (AST) methodology to ensure a thorough and accurate analysis.

\section{Results and Discussion}
This section presents and analyzes the performance of a diverse set of publicly available, instruction-tuned Small Language Models (SLMs) and Large Language Models (LLMs) on our newly developed Hindi benchmark suite, with detailed results presented in Table \ref{tab:hindi_model_benchmarks}. The models evaluated include representatives from prominent families such as Google's Gemma, Meta's Llama, OpenAI's GPT-OSS, NVIDIA's Nemotron, Qwen, and Sarvam, alongside other notable multilingual models.

Among the SLMs, the results reveal a competitive landscape. Gemma-2-9b-it provides the best all-around performance, securing the highest scores on MT-Bench-Hi, IFEval-Hi, and ChatRAG-Hi. Aya-expanse-8b secures the best score on GSM8K-Hi. The value of targeted, language-specific training is highlighted by Nemotron-4-Mini-Hindi-4B-Instruct, which leads significantly on BFCL-Hi.

In the LLM category (models with > 20B parameters), GPT-OSS-120B demonstrates standout performance by achieving the best scores on MT-Bench-Hi, GSM8K-Hi, and IFEval-Hi. Other models show specialized strengths: Gemma-3-27b-it achieves the highest score on BFCL-Hi, while the largest model, Llama-3.1-405B, excels on ChatRAG-Hi. However, it is worth noting that GPT-OSS may have an inherent advantage due to its reasoning mode, even though we set it to a low level for a fairer comparison, and the potential for the GPT-4o judge to be biased towards a sibling OpenAI model also warrants further investigation.

Furthermore, activating the dedicated reasoning modes in models like GPT-OSS and Sarvam-M provides a substantial performance uplift on complex tasks like BFCL, GSM8K, and IFEval. With these capabilities enabled, GPT-OSS-120B achieves the top scores across multiple benchmarks, highlighting the value of reasoning models for Hindi.

In summary, while specialized models show strength in specific tasks, Gemma-2-9b-it in the SLM class and GPT-OSS-120B in the LLM class emerge as the most capable general-purpose models. The distribution of top scores across different models highlights that no single model is best for all tasks. This analysis also indicates that model size is not the sole determinant of performance, a point reinforced by both the 8B Aya model outperforming larger SLMs on GSM8K-Hi and the competitive results of Sarvam-M, which was post-trained on Indic languages. These findings suggest that architectural choices and targeted training data are crucial for developing specialized capabilities for the Hindi language.

\section{Conclusion}
In this work, we addressed the critical gap in evaluation resources for instruction-tuned Hindi LLMs by introducing a new suite of five culturally and linguistically robust benchmarks. Our hybrid curation methodology, combining careful human-centric creation with a translate-and-verify process, provides a valuable framework for developing similar resources in other languages. Our evaluation of various public LLMs supporting the Hindi language revealed a competitive landscape where different models exhibit specialized strengths in reasoning, conversation, and function calling. This suite enables a more nuanced assessment of Hindi LLMs, supporting the broader goal of fostering more equitable and capable multilingual AI systems.

\section*{Limitations}
We acknowledge certain limitations in our work. While our benchmark suite is comprehensive, it does not encompass every possible instruction type or conversational scenario. The use of an "LLM-as-a-Judge" for MT-Bench-Hi carries inherent biases, particularly as the judge model's proficiency in evaluating nuanced Hindi content is not guaranteed. Furthermore, datasets developed through translation, despite human verification, could be improved with full human curation to better capture cultural and linguistic subtleties. Future work could expand the scope of these benchmarks and explore alternative evaluation methodologies.

\section*{Acknowledgements}
This work would not have been possible without contributions from many people at NVIDIA. To mention a few: Priyanka Chavan, Shweta Dash, Asmita Kadwe, Nilesh Chauhan, Saloni Pipada, Shrutika Marke, Priyanka Patil, Ninad Nemade, Muskan Grewal, Aditya Patil, and Noopur Mishra.

\bibliography{main}

\begin{thebibliography}{28}
\providecommand{\natexlab}[1]{#1}

\bibitem[{Achiam et~al.(2023)Achiam, Adler, Agarwal, Ahmad, Akkaya, Aleman, Almeida, Altenschmidt, Altman, Anadkat et~al.}]{achiam2023gpt}
Josh Achiam, Steven Adler, Sandhini Agarwal, Lama Ahmad, Ilge Akkaya, Florencia~Leoni Aleman, Diogo Almeida, Janko Altenschmidt, Sam Altman, Shyamal Anadkat, et~al. 2023.
\newblock Gpt-4 technical report.
\newblock \emph{arXiv preprint arXiv:2303.08774}.

\bibitem[{Austin et~al.(2021)Austin, Odena, Nye, Bosma, Michalewski, Dohan, Jiang, Cai, Terry, Le et~al.}]{austin2021program}
Jacob Austin, Augustus Odena, Maxwell Nye, Maarten Bosma, Henryk Michalewski, David Dohan, Ellen Jiang, Carrie Cai, Michael Terry, Quoc Le, et~al. 2021.
\newblock Program synthesis with large language models.
\newblock \emph{arXiv preprint arXiv:2108.07732}.

\bibitem[{Bender et~al.(2021)Bender, Gebru, McMillan-Major, and Shmitchell}]{bender2021dangers}
Emily~M Bender, Timnit Gebru, Angelina McMillan-Major, and Shmargaret Shmitchell. 2021.
\newblock On the dangers of stochastic parrots: Can language models be too big?
\newblock In \emph{Proceedings of the 2021 ACM conference on fairness, accountability, and transparency}, pages 610--623.

\bibitem[{Chen et~al.(2021)Chen, Tworek, Jun, Yuan, Pinto, Kaplan, Edwards, Burda, Joseph, Brockman et~al.}]{chen2021evaluating}
Mark Chen, Jerry Tworek, Heewoo Jun, Qiming Yuan, Henrique Ponde De~Oliveira Pinto, Jared Kaplan, Harri Edwards, Yuri Burda, Nicholas Joseph, Greg Brockman, et~al. 2021.
\newblock Evaluating large language models trained on code.
\newblock \emph{arXiv preprint arXiv:2107.03374}.

\bibitem[{Cobbe et~al.(2021)Cobbe, Kosaraju, Bavarian, Chen, Jun, Kaiser, Plappert, Tworek, Hilton, Nakano et~al.}]{cobbe2021training}
Karl Cobbe, Vineet Kosaraju, Mohammad Bavarian, Mark Chen, Heewoo Jun, Lukasz Kaiser, Matthias Plappert, Jerry Tworek, Jacob Hilton, Reiichiro Nakano, et~al. 2021.
\newblock Training verifiers to solve math word problems.
\newblock \emph{arXiv preprint arXiv:2110.14168}.

\bibitem[{Doddapaneni et~al.(2023)Doddapaneni, Aralikatte, Ramesh, Goyal, Khapra, Kunchukuttan, and Kumar}]{doddapaneni2023towards}
Sumanth Doddapaneni, Rahul Aralikatte, Gowtham Ramesh, Shreya Goyal, Mitesh~M Khapra, Anoop Kunchukuttan, and Pratyush Kumar. 2023.
\newblock Towards leaving no indic language behind: Building monolingual corpora, benchmark and models for indic languages.
\newblock In \emph{Proceedings of the 61st Annual Meeting of the Association for Computational Linguistics (Volume 1: Long Papers)}, pages 12402--12426.

\bibitem[{Endait et~al.(2025)Endait, Ghatage, Kulkarni, Patil, and Joshi}]{endait2025indicsquad}
Sharvi Endait, Ruturaj Ghatage, Aditya Kulkarni, Rajlaxmi Patil, and Raviraj Joshi. 2025.
\newblock Indicsquad: A comprehensive multilingual question answering dataset for indic languages.
\newblock \emph{arXiv preprint arXiv:2505.03688}.

\bibitem[{Hendrycks et~al.(2020)Hendrycks, Burns, Basart, Zou, Mazeika, Song, and Steinhardt}]{hendrycks2020measuring}
Dan Hendrycks, Collin Burns, Steven Basart, Andy Zou, Mantas Mazeika, Dawn Song, and Jacob Steinhardt. 2020.
\newblock Measuring massive multitask language understanding.
\newblock \emph{arXiv preprint arXiv:2009.03300}.

\bibitem[{Joshi et~al.(2024)Joshi, Singla, Kamath, Kalani, Paul, Vaidya, Chauhan, Wartikar, and Long}]{joshi2024adapting}
Raviraj Joshi, Kanishk Singla, Anusha Kamath, Raunak Kalani, Rakesh Paul, Utkarsh Vaidya, Sanjay~Singh Chauhan, Niranjan Wartikar, and Eileen Long. 2024.
\newblock Adapting multilingual llms to low-resource languages using continued pre-training and synthetic corpus.
\newblock \emph{arXiv preprint arXiv:2410.14815}.

\bibitem[{Kakwani et~al.(2020)Kakwani, Kunchukuttan, Golla, NC, Bhattacharyya, Khapra, and Kumar}]{kakwani2020indicnlpsuite}
Divyanshu Kakwani, Anoop Kunchukuttan, Satish Golla, Gokul NC, Avik Bhattacharyya, Mitesh~M Khapra, and Pratyush Kumar. 2020.
\newblock Indicnlpsuite: Monolingual corpora, evaluation benchmarks and pre-trained multilingual language models for indian languages.
\newblock In \emph{Findings of the association for computational linguistics: EMNLP 2020}, pages 4948--4961.

\bibitem[{Lai et~al.(2023)Lai, Nguyen, Ngo, Nguyen, Dernoncourt, Rossi, and Nguyen}]{lai2023okapi}
Viet Lai, Chien Nguyen, Nghia Ngo, Thuat Nguyen, Franck Dernoncourt, Ryan Rossi, and Thien Nguyen. 2023.
\newblock Okapi: Instruction-tuned large language models in multiple languages with reinforcement learning from human feedback.
\newblock In \emph{Proceedings of the 2023 Conference on Empirical Methods in Natural Language Processing: System Demonstrations}, pages 318--327.

\bibitem[{Liang et~al.(2022)Liang, Bommasani, Lee, Tsipras, Soylu, Yasunaga, Zhang, Narayanan, Wu, Kumar et~al.}]{liang2022holistic}
Percy Liang, Rishi Bommasani, Tony Lee, Dimitris Tsipras, Dilara Soylu, Michihiro Yasunaga, Yian Zhang, Deepak Narayanan, Yuhuai Wu, Ananya Kumar, et~al. 2022.
\newblock Holistic evaluation of language models.
\newblock \emph{arXiv preprint arXiv:2211.09110}.

\bibitem[{Lin et~al.(2022)Lin, Hilton, and Evans}]{lin2022truthfulqa}
Stephanie Lin, Jacob Hilton, and Owain Evans. 2022.
\newblock Truthfulqa: Measuring how models mimic human falsehoods.
\newblock In \emph{Proceedings of the 60th Annual Meeting of the Association for Computational Linguistics (Volume 1: Long Papers)}, pages 3214--3252.

\bibitem[{Patil et~al.()Patil, Mao, Yan, Ji, Suresh, Stoica, and Gonzalez}]{patilberkeley}
Shishir~G Patil, Huanzhi Mao, Fanjia Yan, Charlie Cheng-Jie Ji, Vishnu Suresh, Ion Stoica, and Joseph~E Gonzalez.
\newblock The berkeley function calling leaderboard (bfcl): From tool use to agentic evaluation of large language models.
\newblock In \emph{Forty-second International Conference on Machine Learning}.

\bibitem[{Rohera et~al.(2024)Rohera, Ginimav, Salunke, Sawant, and Joshi}]{rohera2024l3cube}
Pritika Rohera, Chaitrali Ginimav, Akanksha Salunke, Gayatri Sawant, and Raviraj Joshi. 2024.
\newblock L3cube-indicquest: A benchmark question answering dataset for evaluating knowledge of llms in indic context.
\newblock In \emph{Proceedings of the 38th Pacific Asia Conference on Language, Information and Computation}, pages 982--988.

\bibitem[{Sakaguchi et~al.(2021)Sakaguchi, Bras, Bhagavatula, and Choi}]{sakaguchi2021winogrande}
Keisuke Sakaguchi, Ronan~Le Bras, Chandra Bhagavatula, and Yejin Choi. 2021.
\newblock Winogrande: An adversarial winograd schema challenge at scale.
\newblock \emph{Communications of the ACM}, 64(9):99--106.

\bibitem[{Sankalp et~al.()Sankalp, Kumar, Balaji, Kotecha, Jain, Chadha, and Bhaduri}]{sankalp2025indicmmlu}
KJ~Sankalp, Ashutosh Kumar, Laxmaan Balaji, Nikunj Kotecha, Vinija Jain, Aman Chadha, and Sreyoshi Bhaduri.
\newblock Indicmmlu-pro: Benchmarking indic large language models on multi-task language understanding.

\bibitem[{Singh et~al.(2025)Singh, Kumar, Murthy, Sen, Mittal, and Ramakrishnan}]{singh2025indic}
Abhishek~Kumar Singh, Vishwajeet Kumar, Rudra Murthy, Jaydeep Sen, Ashish Mittal, and Ganesh Ramakrishnan. 2025.
\newblock Indic qa benchmark: A multilingual benchmark to evaluate question answering capability of llms for indic languages.
\newblock In \emph{Findings of the Association for Computational Linguistics: NAACL 2025}, pages 2607--2626.

\bibitem[{Singh et~al.(2024{\natexlab{a}})Singh, Gupta, Bharadwaj, Tewari, and Talukdar}]{singh2024indicgenbench}
Harman Singh, Nitish Gupta, Shikhar Bharadwaj, Dinesh Tewari, and Partha Talukdar. 2024{\natexlab{a}}.
\newblock Indicgenbench: A multilingual benchmark to evaluate generation capabilities of llms on indic languages.
\newblock In \emph{Proceedings of the 62nd Annual Meeting of the Association for Computational Linguistics (Volume 1: Long Papers)}, pages 11047--11073.

\bibitem[{Singh et~al.(2024{\natexlab{b}})Singh, Romanou, Fourrier, Adelani, Ngui, Vila-Suero, Limkonchotiwat, Marchisio, Leong, Susanto et~al.}]{singh2024global}
Shivalika Singh, Angelika Romanou, Cl{\'e}mentine Fourrier, David~I Adelani, Jian~Gang Ngui, Daniel Vila-Suero, Peerat Limkonchotiwat, Kelly Marchisio, Wei~Qi Leong, Yosephine Susanto, et~al. 2024{\natexlab{b}}.
\newblock Global mmlu: Understanding and addressing cultural and linguistic biases in multilingual evaluation.
\newblock \emph{arXiv preprint arXiv:2412.03304}.

\bibitem[{Singh et~al.(2024{\natexlab{c}})Singh, Vargus, Dsouza, Karlsson, Mahendiran, Ko, Shandilya, Patel, Mataciunas, OMahony et~al.}]{singh2024aya}
Shivalika Singh, Freddie Vargus, Daniel Dsouza, B{\"o}rje~F Karlsson, Abinaya Mahendiran, Wei-Yin Ko, Herumb Shandilya, Jay Patel, Deividas Mataciunas, Laura OMahony, et~al. 2024{\natexlab{c}}.
\newblock Aya dataset: An open-access collection for multilingual instruction tuning.
\newblock \emph{arXiv preprint arXiv:2402.06619}.

\bibitem[{Srivastava et~al.(2023)Srivastava, Rastogi, Rao, Shoeb, Abid, Fisch, Brown, Santoro, Gupta, Garriga-Alonso et~al.}]{srivastava2023beyond}
Aarohi Srivastava, Abhinav Rastogi, Abhishek Rao, Abu~Awal Shoeb, Abubakar Abid, Adam Fisch, Adam~R Brown, Adam Santoro, Aditya Gupta, Adri Garriga-Alonso, et~al. 2023.
\newblock Beyond the imitation game: Quantifying and extrapolating the capabilities of language models.
\newblock \emph{Transactions on machine learning research}.

\bibitem[{Verma et~al.(2025)Verma, Khan, Kumar, Murthy, and Sen}]{verma2025milu}
Sshubam Verma, Mohammed Safi Ur~Rahman Khan, Vishwajeet Kumar, Rudra Murthy, and Jaydeep Sen. 2025.
\newblock Milu: A multi-task indic language understanding benchmark.
\newblock In \emph{Proceedings of the 2025 Conference of the Nations of the Americas Chapter of the Association for Computational Linguistics: Human Language Technologies (Volume 1: Long Papers)}, pages 10076--10132.

\bibitem[{Wang et~al.(2023)Wang, Chen, Pei, Xie, Kang, Zhang, Xu, Xiong, Dutta, Schaeffer et~al.}]{wang2023decodingtrust}
Boxin Wang, Weixin Chen, Hengzhi Pei, Chulin Xie, Mintong Kang, Chenhui Zhang, Chejian Xu, Zidi Xiong, Ritik Dutta, Rylan Schaeffer, et~al. 2023.
\newblock Decodingtrust: A comprehensive assessment of trustworthiness in gpt models.
\newblock In \emph{NeurIPS}.

\bibitem[{Wang et~al.(2024)Wang, Ma, Zhang, Ni, Chandra, Guo, Ren, Arulraj, He, Jiang et~al.}]{wang2024mmlu}
Yubo Wang, Xueguang Ma, Ge~Zhang, Yuansheng Ni, Abhranil Chandra, Shiguang Guo, Weiming Ren, Aaran Arulraj, Xuan He, Ziyan Jiang, et~al. 2024.
\newblock Mmlu-pro: A more robust and challenging multi-task language understanding benchmark.
\newblock \emph{Advances in Neural Information Processing Systems}, 37:95266--95290.

\bibitem[{Zellers et~al.(2019)Zellers, Holtzman, Bisk, Farhadi, and Choi}]{zellers2019hellaswag}
Rowan Zellers, Ari Holtzman, Yonatan Bisk, Ali Farhadi, and Yejin Choi. 2019.
\newblock Hellaswag: Can a machine really finish your sentence?
\newblock In \emph{Proceedings of the 57th Annual Meeting of the Association for Computational Linguistics}, pages 4791--4800.

\bibitem[{Zheng et~al.(2023)Zheng, Chiang, Sheng, Zhuang, Wu, Zhuang, Lin, Li, Li, Xing et~al.}]{zheng2023judging}
Lianmin Zheng, Wei-Lin Chiang, Ying Sheng, Siyuan Zhuang, Zhanghao Wu, Yonghao Zhuang, Zi~Lin, Zhuohan Li, Dacheng Li, Eric Xing, et~al. 2023.
\newblock Judging llm-as-a-judge with mt-bench and chatbot arena.
\newblock \emph{Advances in Neural Information Processing Systems}, 36:46595--46623.

\bibitem[{Zhou et~al.(2023)Zhou, Lu, Mishra, Brahma, Basu, Luan, Zhou, and Hou}]{zhou2023instruction}
Jeffrey Zhou, Tianjian Lu, Swaroop Mishra, Siddhartha Brahma, Sujoy Basu, Yi~Luan, Denny Zhou, and Le~Hou. 2023.
\newblock Instruction-following evaluation for large language models.
\newblock \emph{arXiv preprint arXiv:2311.07911}.

\end{thebibliography}

\appendix
\section{Appendix}
\label{sec:appendix}
\subsection{Annotator Team and Process}
The human-centric curation and verification tasks were conducted by a team of five specialists. These individuals were employees of our organization, compensated fairly for their work, and were selected for their proficiency in Hindi as either a first or second language. Representing various regions across India, they possessed strong reading and writing skills and a solid understanding of cultural nuances.

The primary tool used for annotation was SuperAnnotate, which provided an intuitive interface for our workflow. This platform allowed for the efficient sharing of examples, processing of results using Python scripts, and performance of quality assurance (QA). The data underwent periodic QA and development checks to ensure alignment with project requirements. To maintain high levels of creativity and productivity, the specialists worked in focused sessions of 2–3 hours per day.

\begin{figure*}[h]  
    \centering
    \includegraphics[scale=0.2]{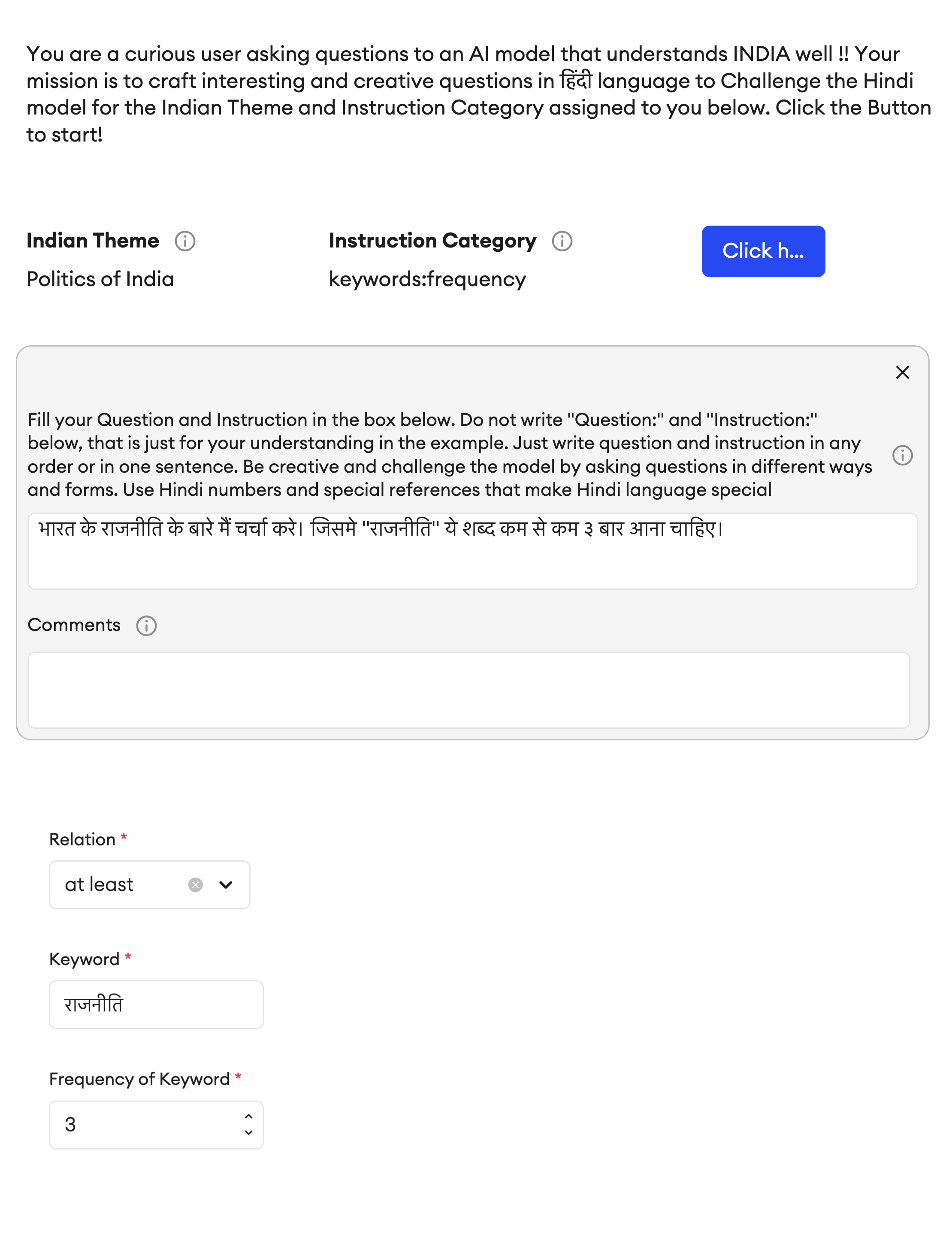} 
    \caption{Illustration of the annotation interface used to curate the IFEval dataset, displaying the guidelines and example instructions provided to annotators.}
    \label{fig:ifeval_ui_1}
\end{figure*}

\begin{figure*}[h]  
    \centering
    \includegraphics[scale=0.5]{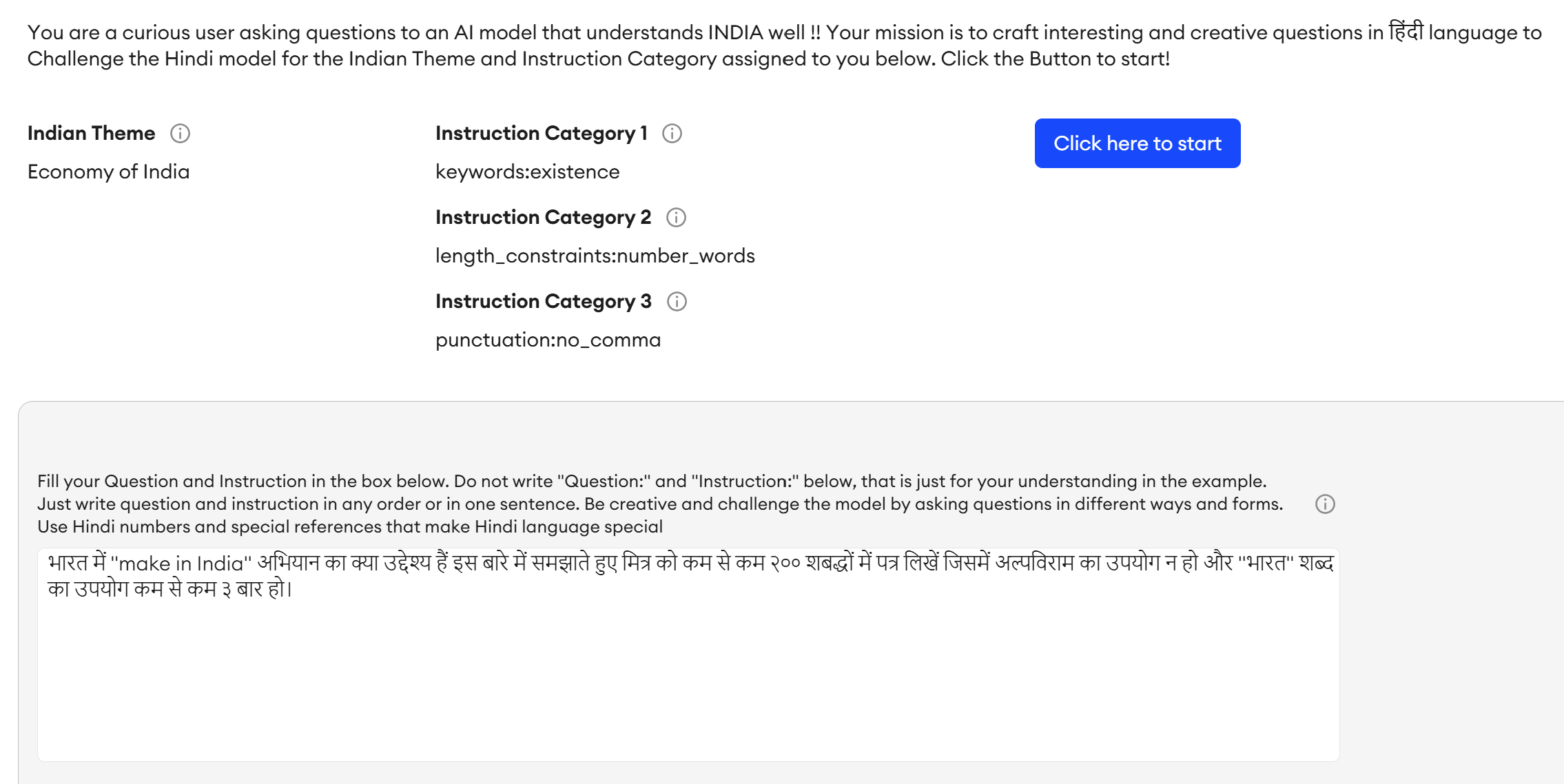}  
    \caption{Example entry from the Hindi IFEval, curated at three levels of complexity: Single Instruction, Double Instruction, and Triple Instruction.}
    \label{fig:ifeval_ui_2}
\end{figure*}
\begin{figure*}[h]  
    \centering
    \includegraphics[scale=0.8]{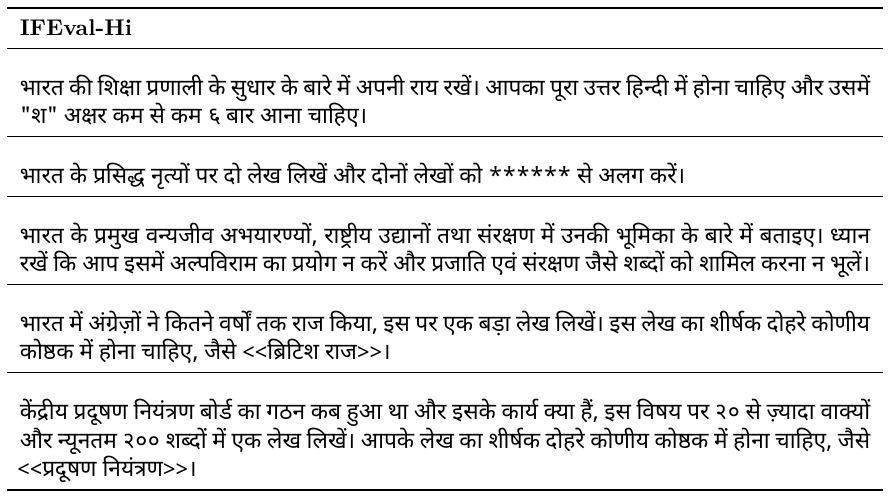}  
    \caption{Examples from the IFEval-Hi benchmark.}
    \label{fig:ifeval_hi_samples}
\end{figure*}

\subsection{IFEval-Hi Curation Process}
\label{subsec:appendix_ifeval}
The annotation procedure for IFEval-Hi was highly structured and communicated to annotators through a comprehensive guidelines document. The process was organized into sequential stages of increasing complexity, beginning with cases that contained a single verifiable instruction, followed by stages with two and then three instructions. Each test case was assigned a predefined Indian theme and instruction category to ensure a balanced distribution of scenarios.

The annotation workflow for each case involved several key components:
\begin{itemize}
    \item Reference Sample: Annotators were provided with a developer-generated sample in Hindi that incorporated the verifiable instruction, serving as a clear reference for the task requirements.
    \item Annotation Interface: A dedicated text box was provided for annotators to formulate their questions based on the assigned theme and instruction category, with a separate comments box for any necessary clarifications with the quality control developer.
    \item Evaluation Parameters: The parameters required for automatic evaluation were also systematically recorded, aligning with the standards of the English dataset. Annotators received detailed guidelines for these parameters, with the Hindi reference sample serving as a practical model.
    \item Review and Feedback Loop: A weekly review of approximately 50\% of submitted samples was conducted by developers. Any cases requiring revision were returned to annotators with specific feedback in the comments section, ensuring a consistent feedback loop and high-quality output.
\end{itemize}
Sample annotation UI screens are shown in Figures \ref{fig:ifeval_ui_1} and \ref{fig:ifeval_ui_2}. Some examples from the dataset are shown in Figure \ref{fig:ifeval_hi_samples}.

\begin{figure*}[h]  
    \centering
    \includegraphics[scale=0.15]{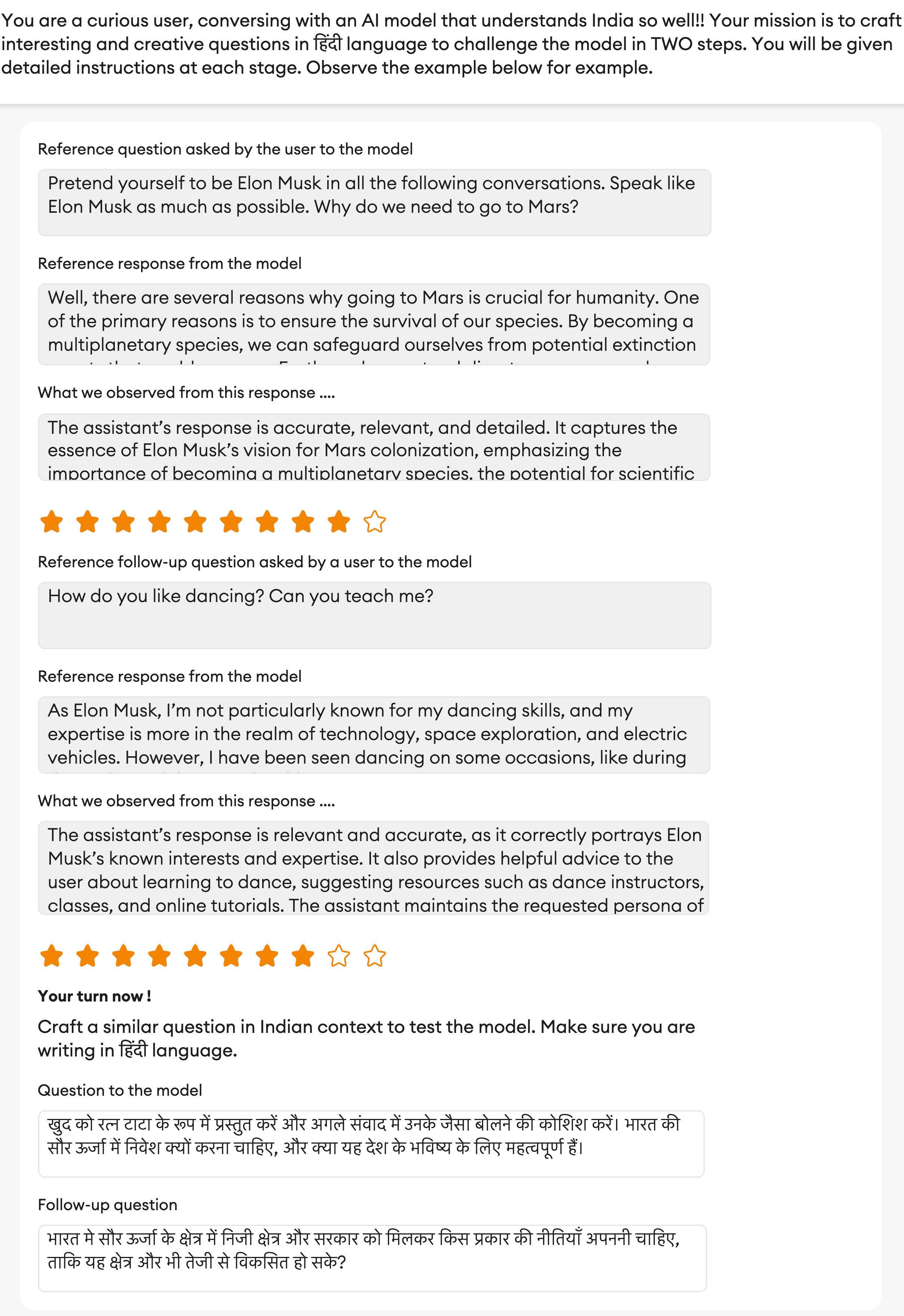}
    \caption{Illustration of the annotation interface used to curate the culturally adapted Indic version of the MT-Bench dataset, displaying the guidelines and example instructions provided to annotators.}
    \label{fig:mtbench_ui}
\end{figure*}

\begin{figure*}[h]  
    \centering 
    \includegraphics[scale=0.9]{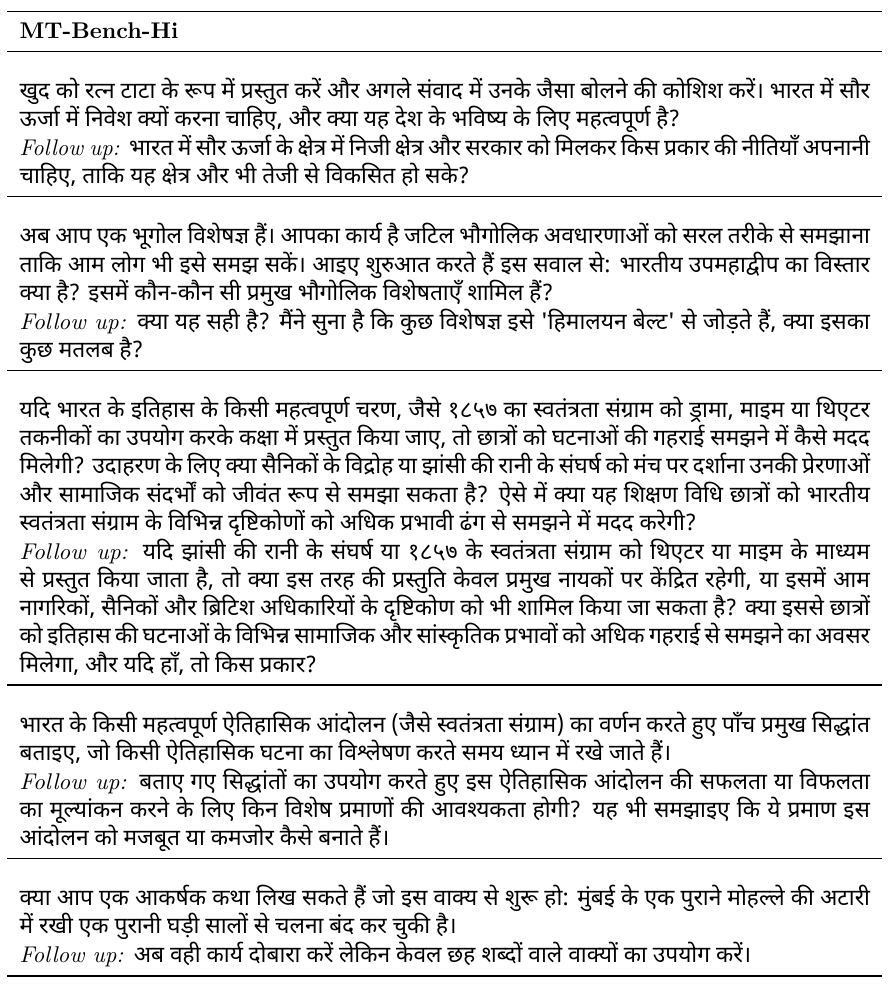}
    \caption{Examples from the MT-Bench-Hi benchmark.}
    \label{fig:mtbench_hi_samples}
\end{figure*}

\subsection{MT-Bench-Hi Curation Process}
\label{subsec:appendix_mtbench}
The curation of the MT-Bench-Hi dataset, while presenting distinct challenges, benefited from the procedural learnings established during the IFEval-Hi creation. Specialists were guided by supplementary instructions tailored to the specific demands of creating multi-turn conversational benchmarks. This process was designed to help annotators understand the evaluation procedure and produce high-quality, contextually relevant samples.

The workflow for each test case provided annotators with a comprehensive view of the task:
\begin{itemize}
    \item Original English Sample: Annotators were given an original question and follow-up from the MT-Bench dataset as a reference.
    \item Model Response Example: The corresponding model-generated response for the English sample was included.
    \item Evaluation Insight: The AI judge's rating and judgment for that response were also provided, offering annotators direct insight into the evaluation criteria and performance expectations.
\end{itemize}
Using this framework, annotators reviewed the initial English question and its follow-up, then crafted analogous questions contextualized for Indian settings. To ensure quality and adherence to guidelines, 50\% of the newly created samples were subject to a weekly review by a developer. This structured approach equipped annotators to produce high-quality, contextually appropriate samples for the MT-Bench-Hi dataset. The sample annotation UI screen is shown in Figure \ref{fig:mtbench_ui}. Some examples from the dataset are shown in Figure \ref{fig:mtbench_hi_samples}.

\begin{figure*}[h]  
    \centering
    \includegraphics[scale=0.2]{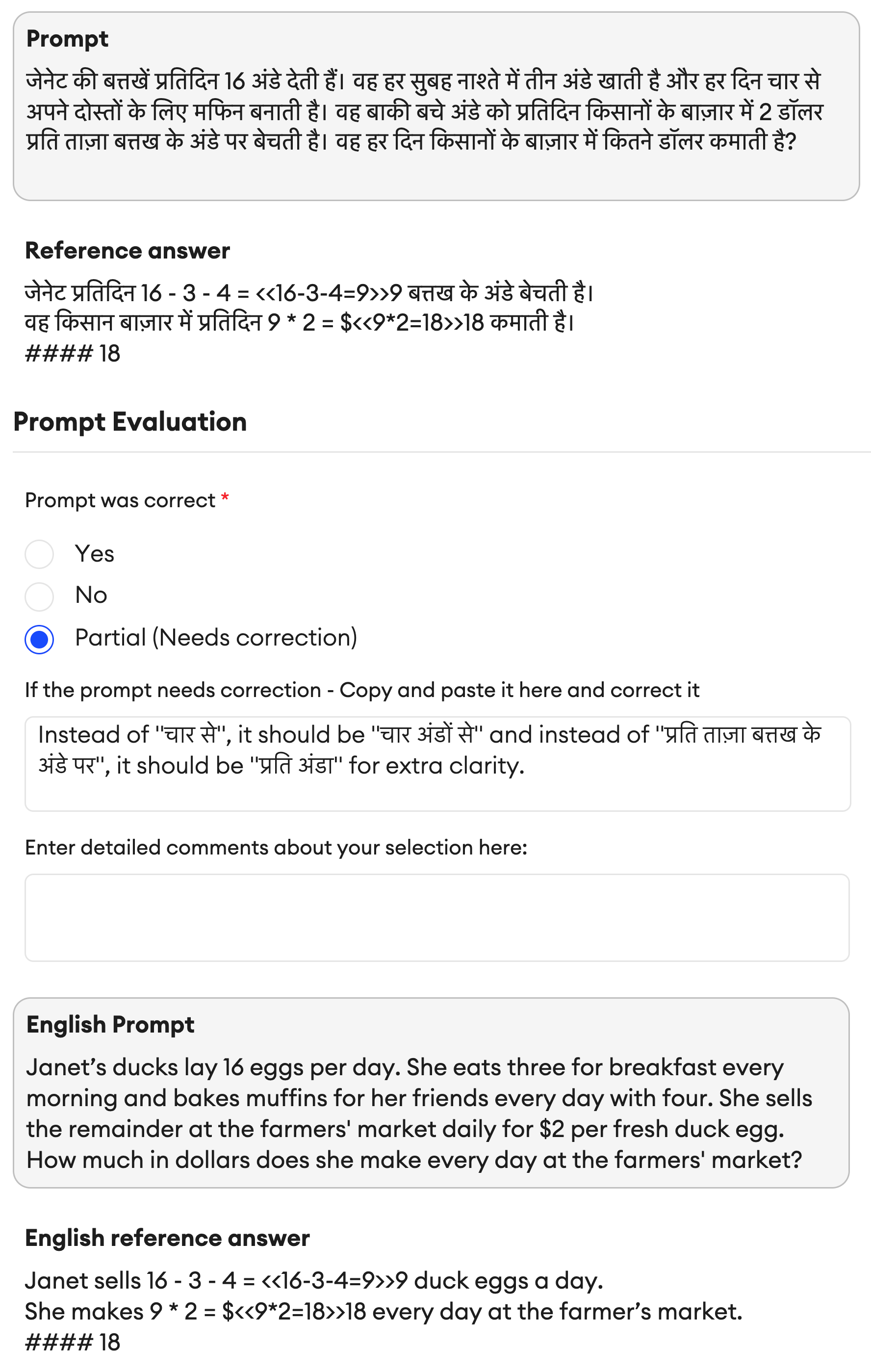}
    \caption{Illustration of the annotation interface used to evaluate the translation quality of the GSM8K dataset, displaying the guidelines and example instructions provided to annotators.}
    \label{fig:gsm8k_ui}
\end{figure*}

\begin{figure*}[h]  
    \centering
    \includegraphics[scale=0.8]{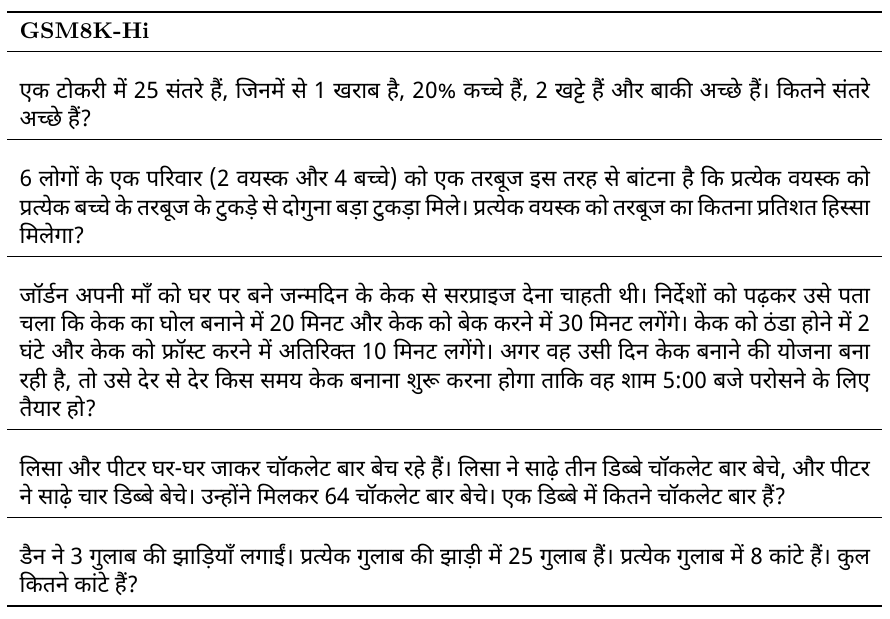}
    \caption{Examples from the GSM8K-Hi benchmark.}
    \label{fig:gsm8k_hi_samples}
\end{figure*}

\begin{figure*}[h]  
    \centering
    \includegraphics[scale=0.8]{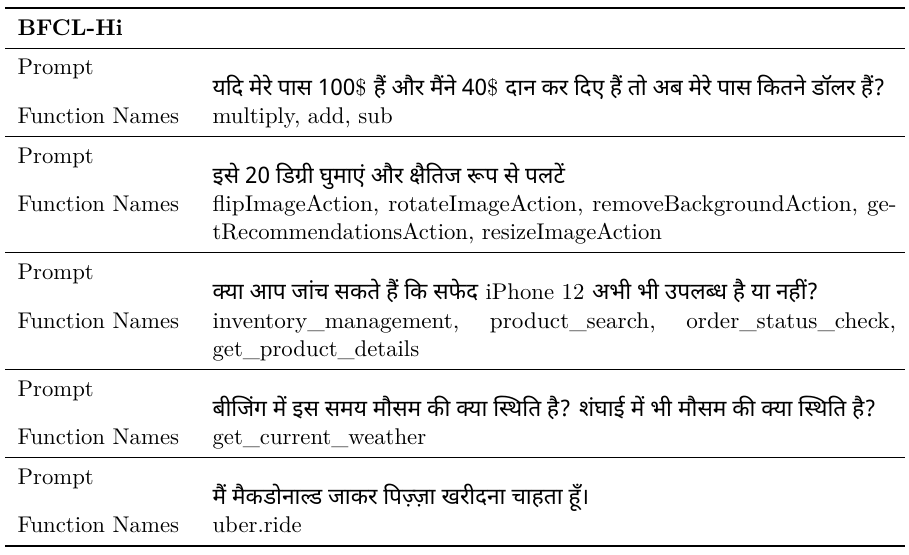}
    \caption{Examples from the BFCL-Hi dataset.}
    \label{fig:bfcl_hi_samples}
\end{figure*}

\subsection{GSM8K-Hi Curation Process}
\label{subsec:appendix_gsm8k}
By the GSM8K annotation stage, annotators were proficient with the annotation interface. The mathematical nature of this dataset required sustained attention to detail during the verification process.

The workflow for each sample test case included the following elements:
\begin{itemize}
    \item Translated Content: Annotators received the translated Hindi version of the instruction and the corresponding translated output.
    \item Final Answer: The final numerical answer was clearly indicated for verification.
\end{itemize}
Annotators were instructed to carefully read and comprehend the question to assess its clarity and coherence, using the provided solution for additional context if necessary. They were tasked with flagging any ambiguous or unclear questions for review. On the quality control interface, developers reviewed the annotated samples, referencing the original English versions to guide any necessary corrections.
The sample annotation UI screen is shown in Figure \ref{fig:gsm8k_ui}. Some examples from the dataset are shown in Figure \ref{fig:gsm8k_hi_samples}.

\subsection{ChatRAG-Hi and BFCL-Hi Curation Process}
\begin{figure*}[h]  
    \centering
    \includegraphics[scale=0.9]{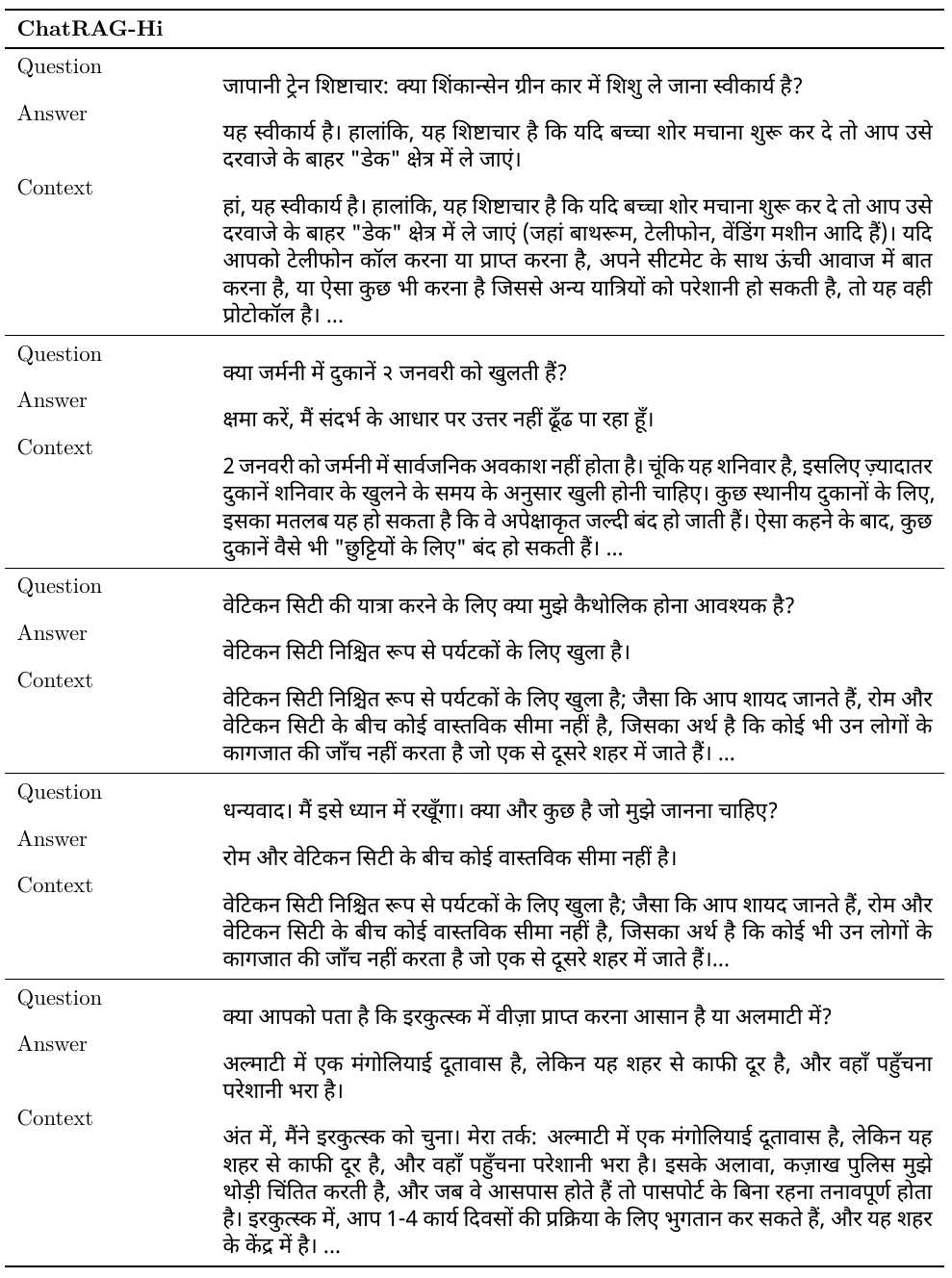}
    \caption{Examples from ChatRAG-Hi dataset. Each example contains a user question, a single answer, and partial supporting context for illustration.}
    \label{fig:chat_rag_hi_samples}
\end{figure*}

The ChatRAG-Hi and BFCL-Hi datasets were curated through GCP translation and subsequent filtering. Examples are provided in Figures \ref{fig:chat_rag_hi_samples} and \ref{fig:bfcl_hi_samples}, respectively.

\end{document}